\documentclass[10pt]{article}

\usepackage[letterpaper, margin=1in, headheight=25pt]{geometry}
\usepackage{times}           
\usepackage{graphicx}        
\usepackage{amsmath}         
\usepackage{cite}            
\usepackage{url}             
\usepackage{fancyhdr}        
\usepackage{titlesec}        
\usepackage{caption}         
\usepackage{parskip}         
\usepackage{algorithm}       
\usepackage{enumitem}        
\usepackage[T1]{fontenc}
\usepackage[utf8]{inputenc}

\setlist[itemize]{noitemsep, topsep=2pt}

\usepackage{etoolbox}        
\usepackage{algpseudocode}
\usepackage{etoolbox}

\patchcmd{\thebibliography}
  {\setlength{\itemsep}{\z@}}
  {\setlength{\itemsep}{0pt}\setlength{\parskip}{0pt}}
  {}{}

\usepackage{placeins}
\usepackage[colorlinks=true,
            linkcolor=black,
            citecolor=black,
            urlcolor=black]{hyperref}

\titleformat{\section}
    {\normalfont\fontsize{12}{14}\bfseries\selectfont}
    {\thesection.}{0.5em}{}
\titlespacing*{\section}{0pt}{6pt}{0pt}

\titleformat{\subsection}
    {\normalfont\fontsize{10}{12}\bfseries\selectfont}
    {\thesubsection}{0.5em}{}
\titlespacing*{\subsection}{0pt}{6pt}{0pt}

\titleformat{\subsubsection}
    {\normalfont\fontsize{10}{12}\bfseries\selectfont}
    {\thesubsubsection}{0.5em}{}
\titlespacing*{\subsubsection}{0pt}{6pt}{0pt}

\setlist[itemize,1]{label=\textbullet, leftmargin=0.5in}
\setlist[itemize,2]{label=\textopenbullet, leftmargin=0.75in}
\setlist[itemize,3]{label=\textendash, leftmargin=1in}

\begin{document}
\thispagestyle{plain}

\begin{center}
{\fontsize{16}{19}\bfseries\selectfont 
Adaptive Multi-Expert Graph Transformer for Interpretable EEG-Based Diagnostics}

\vspace{10pt}

{\fontsize{10}{12}\selectfont
Maryam Rahimimovassagh, Md Elias Hossain, Ivan Garibay, Waldemar Karwowski, Niloofar Yousefi \\

\vspace{5pt}

Department of Industrial Engineering and Management Systems \\
University of Central Florida, Orlando, FL, USA \\

\vspace{5pt}

\texttt{maryam.rahimimovassagh@ucf.edu, mdelias.hossain@ucf.edu, igaribay@ucf.edu, wkar@ucf.edu, niloofar.yousefi@ucf.edu}
}
\end{center}

\vspace{10pt}

\vspace{10pt}

\begin{center}
{\fontsize{12}{14}\bfseries\selectfont Abstract}
\end{center}

{\fontsize{10}{12}\selectfont
 Electroencephalographic (EEG) abnormalities arise from dynamic changes in neural
synchrony across spatial and temporal scales, yet many computational approaches
reduce these dynamics to static features. We present a Spatial Multi-Expert Graph
Transformer that models each EEG recording as a sequence of dynamic functional
connectivity graphs. Time-resolved connectivity is estimated using the weighted
Phase Lag Index (wPLI), and hierarchical graph encoding aggregates information from
electrode to regional and global levels. A multi-expert transformer architecture
enables subtype-aware reasoning, with a gating mechanism adaptively fusing expert
outputs for global abnormality prediction. Experiments on the TUAB dataset show competitive abnormal EEG detection performance and demonstrate the potential of dynamic graph modeling with adaptive expert fusion for interpretable, subtype-aware spatial--temporal analysis.

}


\section{Introduction}

Electroencephalography (EEG) \cite{blinowska2006electroencephalography} is a widely used
non-invasive technique for assessing brain activity in clinical and research settings.
EEG plays a critical role in detecting neurological abnormalities such as epileptiform
discharges \cite{sanchez2015should}, focal slowing, diffuse slowing, and hemispheric
asymmetries. EEG interpretation relies on identifying spatially localized and temporally evolving patterns, rather than treating recordings as static signals. However, many automated EEG analysis methods still assign a single global label
to an entire recording, limiting their ability to reflect clinically meaningful dynamics.

Recent deep learning approaches have improved EEG abnormality detection using
temporal--spectral modeling, transformer-based architectures, and graph-based
representations. CNN--LSTM hybrids and spectral networks such as FFCL \cite{li2022ffcl}
and SPaRCNet \cite{jing2023sparcnet} capture short-range temporal dynamics, while
transformer-based models model long-range dependencies, including S3T \cite{song2021s3t},
BIOT \cite{yang2024biot}, and LaBraM \cite{jiang2024labrambase}. Graph-based methods
further encode spatial structure through correlation-aware and local--global connectivity
modeling, such as Corr-DCRNN \cite{tang2022corrdcrnn}, LGGNet \cite{ding2024lggnet}, and
XAIguiFormer \cite{guo2025xaiguiformer}. Despite these advances, most existing methods rely
on static connectivity representations and single-expert reasoning, limiting their
ability to capture dynamic and subtype-specific EEG abnormalities.

To address these limitations, this work introduces a Spatial Multi-Expert Graph
Transformer that models EEG recordings as sequences of dynamic functional connectivity
graphs. Functional connectivity is estimated using the wPLI
\cite{hardmeier2014reproducibility}, enabling time-resolved graph representations of
evolving neural synchrony. A hierarchical graph encoder aggregates information from
electrode-level features to regional and global representations, while a multi-expert
transformer enables subtype-aware reasoning through adaptive expert gating. This design is intended to support robustness, interpretability, and alignment with clinical EEG practice.


\section{Framework Overview}

\begin{figure}[t]
\centering
\includegraphics[width=0.85\linewidth]{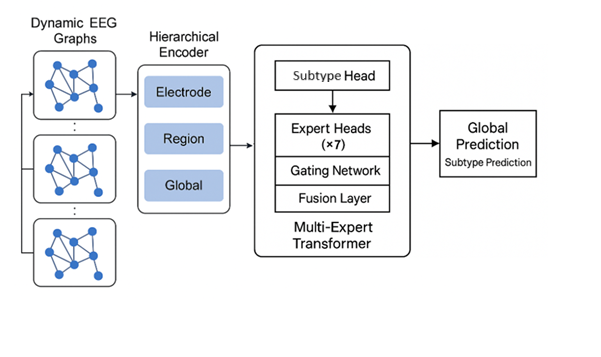}
\caption{Pipeline of dynamic EEG graph construction, hierarchical encoding, and the multi-expert transformer for subtype-aware and global EEG abnormality detection.
}
\label{fig:pipeline}
\end{figure}

The proposed framework transforms EEG recordings into sequences of dynamic functional connectivity graphs for subtype-aware abnormality detection through hierarchical graph encoding and a multi-expert temporal transformer (Figure~\ref{fig:pipeline}). The model employs a three-tier spatial hierarchy, progressing from electrode-level encoding to regional aggregation and global fusion for classification. Figure~\ref{fig:connectivity_heatmaps} illustrates connectivity aggregation from individual electrodes to regional and global representations.

\begin{figure*}[!htbp]
\centering
\includegraphics[width=\linewidth]{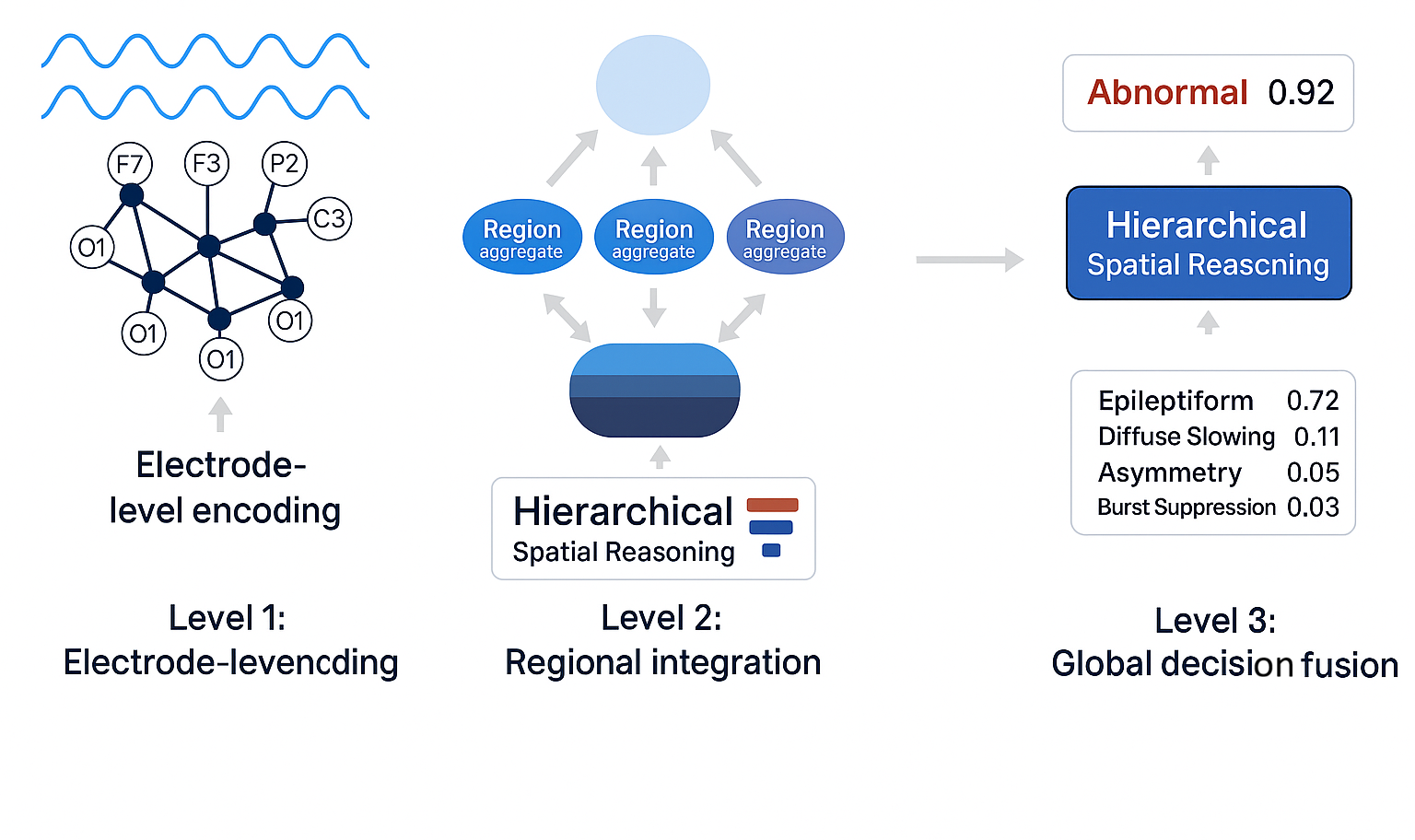}
\caption{Hierarchical spatial encoding and multi-expert temporal reasoning framework.}
\label{fig:framework}

\end{figure*}

EEG recordings are preprocessed using a 1--40~Hz bandpass filter, artifact attenuation, and resampling to 250~Hz. Each recording is divided into overlapping temporal windows, with each window converted into a functional connectivity graph using the wPLI. Nodes represent electrodes and weighted edges encode pairwise interactions, producing a time-ordered sequence of dynamic brain graphs. A hierarchical graph encoder processes each graph at three resolutions: electrode-level graph convolutions extract local features, regional pooling models meso-scale interactions, and global aggregation summarizes brain-wide connectivity patterns.

The encoded graph sequence is processed by a temporal transformer modeling long-range dependencies. A multi-expert module with specialized heads for distinct abnormal subtypes receives transformer outputs, with a learned gating network assigning adaptive expert weights based on input representations. Fused expert outputs generate subtype-specific probabilities and global abnormality predictions, enabling interpretable reasoning with subtype-specific contributions.

Algorithm~\ref{algo_smegt} summarizes the end-to-end pipeline, detailing dynamic graph construction,
hierarchical encoding, temporal modeling, and gated multi-expert decision fusion.

\begin{algorithm}[!t]
\caption{Spatial Multi-Expert Graph Transformer for Dynamic EEG Abnormality Detection}
\label{algo_smegt}
\begin{algorithmic}[1]

\Require EEG recording $X$, number of experts $K$, number of temporal segments $S$
\Ensure Global abnormality label $\hat{y}$ and subtype probabilities $p_{\text{sub}}$

\State Preprocess EEG signal and obtain matrix $X$

\Statex
\State \textbf{Dynamic Graph Construction}
\State Split $X$ into $S$ temporal segments
\For{each segment $W_t$}
    \State Compute functional connectivity $A_t$ using wPLI
   \State Compute channel features $F_t \Leftarrow [\text{mean}(W_t),\,\text{std}(W_t),\,\text{ptp}(W_t)]$

    \State Construct graph frame $G_t = (\mathcal{V}, A_t, F_t)$
\EndFor

\Statex
\State \textbf{Hierarchical Graph Encoding}
\For{each graph $G_t$}
    \State Encode graph to embedding $h_t$
\EndFor

\Statex
\State \textbf{Temporal Modeling}
\State Apply TransformerEncoder to $\{h_t\}_{t=1}^{S}$ to obtain sequence representation $h_{\text{seq}}$

\Statex
\State \textbf{Multi-Expert Reasoning}
\State Compute expert weights $g = \text{softmax}(W_g h_{\text{seq}})$
\For{each expert $k = 1,\dots,K$}
    \State Compute expert output $z_k = \text{Expert}_k(h_{\text{seq}})$
\EndFor
\State Fuse experts: $z_{\text{fused}} = \sum_{k=1}^{K} g_k z_k$

\Statex
\State \textbf{Prediction}
\State Compute subtype probabilities $p_{\text{sub}} = \text{softmax}(W_{\text{sub}} h_{\text{seq}})$
\State $\hat{y} \Leftarrow \arg\max(z_{\text{fused}})$

\end{algorithmic}
\end{algorithm}

\subsection{Implementation Details}

\textbf{Dataset and splits.}
All experiments were conducted on the Temple University Hospital (TUH) EEG corpus~\cite{obeid2016tuh}, a large-scale clinically curated dataset that has been widely adopted for abnormal EEG detection and seizure analysis~\cite{li2022ffcl,jing2023sparcnet,ding2024lggnet}. Data were
split into training (80\%) and test (20\%) sets with stratification by class.
Splits were performed at the recording level; patient-level metadata were not
available.

\textbf{Windowing and graph construction.}
Each EEG recording was resampled to 250 Hz and segmented into $S=6$ non-overlapping
temporal windows of equal length. For each window, a functional connectivity
graph was constructed using wPLI in the 4–30 Hz band.
When wPLI computation was unstable, Pearson correlation was used as a fallback.
Graphs were fully connected with self-loops removed.

\textbf{Node features.}
Each node (electrode) was represented using three normalized channel-wise
statistics computed per window: mean amplitude, standard deviation, and
peak-to-peak range. Hjorth parameters and entropy were used only for preliminary
analysis and not as model inputs.

\textbf{Model configuration.}
The hierarchical graph encoder used three GCN layers with hidden dimension 128,
interleaved with self-attention graph pooling (SAGPool). The temporal module
consisted of a 2-layer transformer encoder with 4 attention heads. The full
model employed $K=7$ subtype-specific experts with a learned gating mechanism.

\textbf{Training.}
Models were trained using AdamW (learning rate $5\times10^{-4}$, weight decay
$10^{-4}$) for 15 epochs. The objective combined global classification loss,
expert loss, subtype prediction loss, and hierarchical consistency regularization.

\section{Results and Analysis}

This section evaluates the proposed Spatial Multi-Expert Graph Transformer,
including baseline performance, hyperparameter sensitivity, optimization
effects, and final model behavior on the TUAB dataset.

\subsection{Preliminary Analyses and Motivation for Spatial Hierarchy}

To evaluate whether EEG abnormalities can be distinguished using non-spatial
features alone, channel-wise temporal descriptors (Hjorth mobility, Hjorth
complexity, and signal entropy) were extracted from 1,500 TUAB recordings for
preliminary analysis and aggregated across channels.
Substantial overlap between normal and abnormal EEGs was observed,
indicating that temporal statistics alone lack sufficient discriminative power.

Each EEG was therefore represented as a functional connectivity graph, where
nodes correspond to electrodes and edges encode phase-based synchrony. Mean
adjacency matrices were computed for each electrographic subtype and compared
against the normal baseline using $\Delta$-connectivity maps:
\begin{equation}
\Delta M_{\text{subtype}} = \overline{A}_{\text{subtype}} - \overline{A}_{\text{normal}}
\end{equation}

\begin{figure*}[!htbp]
\centering
\includegraphics[width=\linewidth]{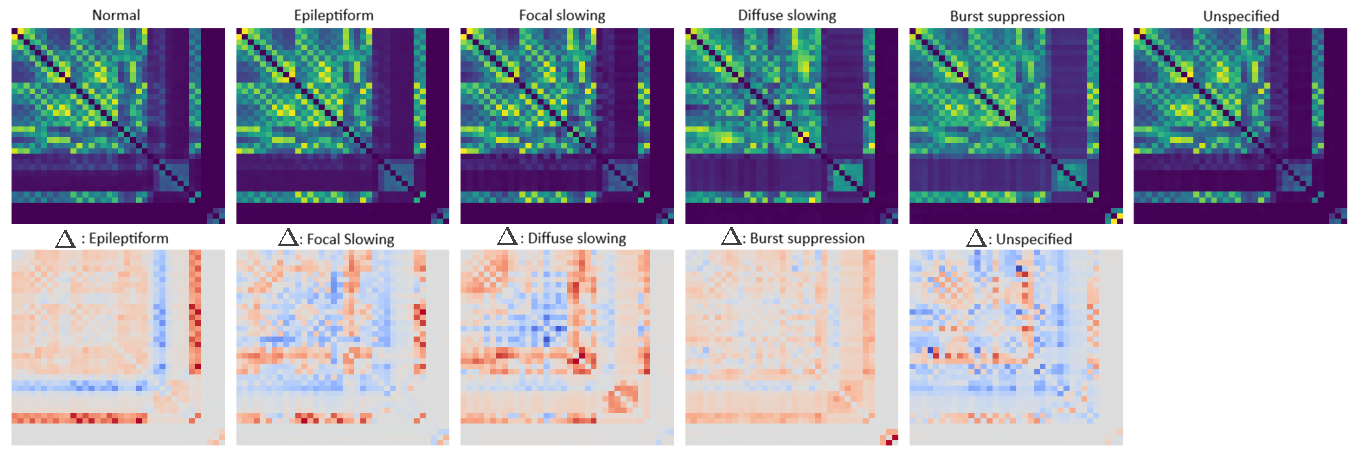}
\caption{Mean connectivity (top) and $\Delta$-connectivity (bottom) maps comparing normal and 
abnormal EEG subtypes. Distinct connectivity fingerprints particularly in focal slowing demonstrate spatially structured deviations from the normal baseline.}
\label{fig:connectivity_heatmaps}
\end{figure*}

As shown in Figure~\ref{fig:connectivity_heatmaps}, distinct subtype-specific
spatial deviations emerge, including localized fronto--temporal
hyperconnectivity, lateralized weakening in focal abnormalities, and global
reductions in long-range connectivity for diffuse slowing. These observations
demonstrate that clinically meaningful EEG abnormalities are primarily encoded
in spatial interaction patterns, motivating hierarchical graph-based modeling.

\subsection{Model Performance and Optimization}

The full gated graph transformer achieved an accuracy of 0.7152, with macro-averaged precision, recall, and F1 score of 0.72, 0.71, and 0.71, respectively. Performance was higher for normal EEGs (recall 0.78) than abnormal EEGs (recall 0.64), revealing a 14-point recall gap consistent with the heterogeneous and transient nature of abnormal EEG patterns.

\FloatBarrier

\begin{table}[t]
\centering
\caption{Ablation study for TUAB abnormal EEG detection using reproducible model variants.}
\label{tab:ablation}
\renewcommand{\arraystretch}{1.15}
\small
\setlength{\tabcolsep}{4pt}
\begin{tabular}{|l|c|c|c|c|c|c|c|c|}
\hline
\textbf{Model Variant} & \textbf{Graph} & \textbf{Hierarchy} & \textbf{Gating} &
\textbf{Acc.} & \textbf{N-P} & \textbf{N-R} & \textbf{A-P} & \textbf{A-R} \\
\hline
A0: No Graph (Feature-Only)
& No & No & Yes
& 0.7027 & 0.67 & 0.79 & 0.74 & 0.60 \\
\hline
A3: Graph + Hierarchy (Single-Expert)
& Yes & Yes & No
& 0.7027 & 0.71 & 0.70 & 0.69 & 0.71 \\
\hline
\textbf{Full Model: Graph + Hierarchy + Gating}
& \textbf{Yes} & \textbf{Yes} & \textbf{Yes}
& \textbf{0.7152} & \textbf{0.69} & \textbf{0.78} & \textbf{0.74} & \textbf{0.64} \\
\hline
\end{tabular}
\end{table}

The reproducible ablation variants focus on feature-only modeling, hierarchical graph encoding, and adaptive expert gating. This allows the analysis to emphasize stable comparisons across the reported model components.

\paragraph{Ablation Analysis.}
Table~\ref{tab:ablation} reports the reproducible ablation variants used to assess the contribution of graph-based modeling and adaptive expert gating. Feature-only modeling (A0) yields competitive accuracy but lower abnormal recall, suggesting that channel-wise temporal statistics alone may not fully capture heterogeneous abnormal patterns. Adding hierarchical graph encoding in a single-expert setting (A3) improves recall balance, indicating that spatial connectivity structure contributes to abnormal-class sensitivity. The full model combines hierarchical graph encoding with adaptive expert gating and achieves the highest overall accuracy while maintaining a balanced precision--recall profile.

\subsubsection{Threshold Sensitivity and Recall--Precision Trade-offs}

While the multi-expert model improves overall performance and abnormal-class
precision, a modest reduction in abnormal recall is observed at the default
decision threshold of 0.5 (Table~\ref{tab:ablation}). To determine whether
this reflects a loss of discriminative capacity or a shift in operating point, we
performed a post-hoc threshold sensitivity analysis without retraining.

\begin{figure*}[!htbp]
\centering
\includegraphics[width=\linewidth]{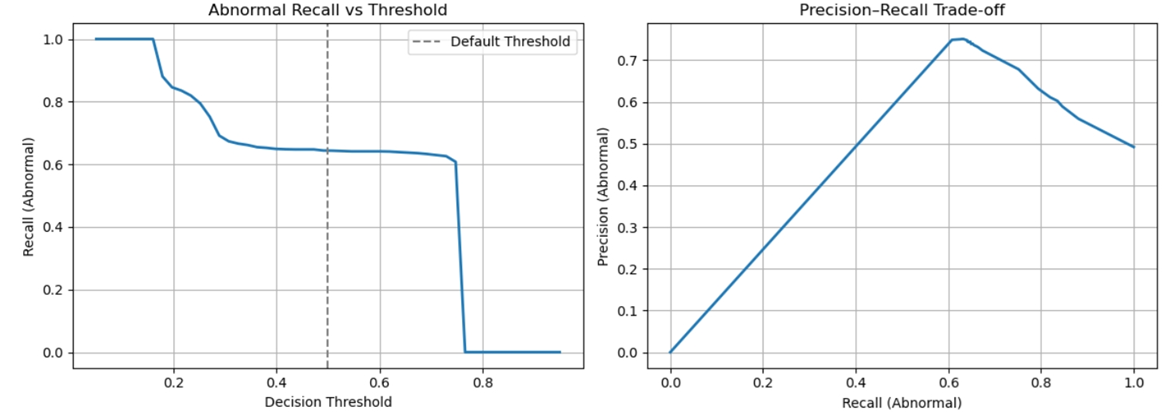}
\caption{Post-hoc threshold analysis of the gated model.
Abnormal recall versus decision threshold (left) and corresponding
precision--recall trade-off (right).}
\label{fig:threshold_analysis}
\end{figure*}

\paragraph{Post-hoc Expert Contribution Analysis.}
Post-hoc analysis of the gating mechanism shows high-entropy expert utilization
under weakly structured inputs, indicating stable routing rather than expert
collapse. In addition, average gating distributions differed across abnormal
subtypes, suggesting emergent expert specialization even without explicit
subtype-wise supervision. Expert-level ablations requiring retraining are
deferred to future work.

\noindent\textbf{Hyperparameter tuning.}
Limited tuning did not yield consistent gains; default settings are used throughout.

\noindent\textbf{Optimization sensitivity.}
We evaluated multiple optimizer, activation, and learning-rate scheduling configurations while fixing the architecture. Performance differences were modest across settings, suggesting that the model remains sensitive to optimization choices and that architectural comparisons should be interpreted with this variability in mind.

\FloatBarrier


\begin{figure}[!htbp]
\centering
\includegraphics[width=0.75\linewidth]{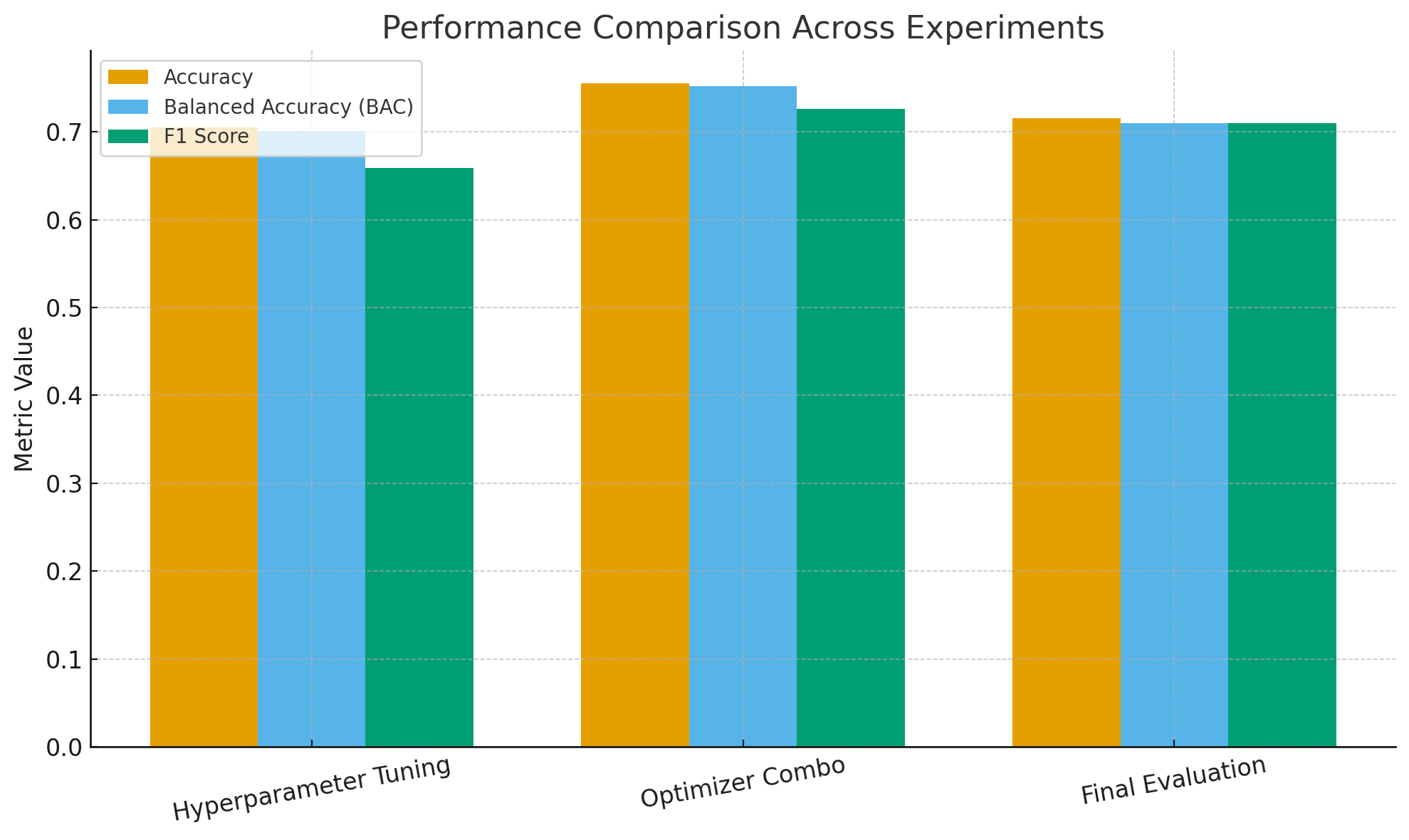}
\caption{Performance comparison across tuning, optimization, and final evaluation stages.}
\label{fig:performance_comparison}
\end{figure}

\subsection{Interpretation}

Overall, the results indicate that representing EEG recordings as dynamic spatial connectivity graphs enables interpretable analysis of abnormal EEG patterns, with distinct electrographic subtypes exhibiting characteristic connectivity differences. The multi-expert transformer facilitates subtype-aware reasoning and adaptive decision fusion, while threshold analysis highlights the importance of operating-point selection for balancing precision and recall.
Observed recall variability reflects clinical heterogeneity and operating-point selection rather than loss of discriminative signal.

\FloatBarrier

\section{Conclusion and Future Work}

This work presents a Spatial Multi-Expert Graph Transformer for EEG abnormality
detection that models EEG signals as dynamic connectivity graphs with hierarchical
encoding for subtype-aware and interpretable predictions. The framework achieves
competitive performance while remaining sensitive to dataset-specific preprocessing
and electrode layout variability.

Future work will focus on incorporating anatomically informed hierarchical pooling to better align graph abstractions with neurophysiological structure, beyond purely data-driven pooling methods. In addition, recall-aware expert gating strategies will be explored to prioritize abnormal-class sensitivity under clinically asymmetric error costs. Extensions toward causal interpretability and multimodal integration will also be investigated.

\section{Competing interests}
The authors declare that there is no competing interest in this study.


\begingroup
\footnotesize
\setlength{\parskip}{0pt}
\setlength{\itemsep}{0pt}
\setlength{\parsep}{0pt}
\setlength{\topsep}{0pt}
\bibliographystyle{ieeetr}
\bibliography{references}

@article{li2022ffcl,
  title={FFCL: Feature Fusion of CNN and LSTM for Abnormal EEG Detection},
  author={Li, Zhi and Zhao, Wei and Xu, Jian and Liu, Peng},
  journal={IEEE Transactions on Neural Systems and Rehabilitation Engineering},
  volume={30},
  pages={1745--1756},
  year={2022},
  publisher={IEEE}
}

@article{jing2023sparcnet,
  title={SPaRCNet: Spectral–Temporal Residual Convolutional Network for EEG Classification},
  author={Jing, Wenhao and Zhou, Kai and Zhang, Lei and Chen, Xiaowei},
  journal={Frontiers in Neuroscience},
  volume={17},
  pages={1208754},
  year={2023},
  publisher={Frontiers}
}

@article{yang2024biot,
  title={BIOT: BioSignal Transformer for Clinical EEG Interpretation},
  author={Yang, Xue and Zhang, Hong and Wang, Qian and Zhang, Zhen},
  journal={Neurocomputing},
  volume={567},
  pages={127018},
  year={2024},
  publisher={Elsevier}
}

@inproceedings{song2021s3t,
  title={S3T: Spatial–Spectral–Temporal Transformer for EEG Classification},
  author={Song, Jiyu and Wang, Cheng and Lu, Zhiyuan and Zhang, Feng},
  booktitle={Proceedings of the IEEE International Conference on Acoustics, Speech and Signal Processing (ICASSP)},
  pages={1234--1238},
  year={2021},
  organization={IEEE}
}

@article{jiang2024labrambase,
  title={LaBraM: Large-Brain Transformer for EEG Foundation Model Pretraining},
  author={Jiang, Haoran and Liu, Qi and Zhang, Yuxuan and He, Wei},
  journal={IEEE Transactions on Biomedical Engineering},
  volume={71},
  number={6},
  pages={1902--1915},
  year={2024},
  publisher={IEEE}
}

@article{tang2022corrdcrnn,
  title={Corr-DCRNN: Correlation-Aware Diffusion Convolutional Recurrent Network for EEG Representation Learning},
  author={Tang, Ruoyu and Zhou, Xin and Wang, Haotian and Chen, Jian},
  journal={Pattern Recognition},
  volume={133},
  pages={108980},
  year={2022},
  publisher={Elsevier}
}

@article{ding2024lggnet,
  title={LGGNet: Local-Global Graph Neural Network for EEG-Based Abnormality Detection},
  author={Ding, Rui and Zhang, Yu and Li, Qiang and He, Tian},
  journal={IEEE Journal of Biomedical and Health Informatics},
  volume={28},
  number={5},
  pages={2101--2114},
  year={2024},
  publisher={IEEE}
}

@article{guo2025xaiguiformer,
  title={XAIguiFormer: Explainable Transformer with Connectome Tokenization for EEG Disorder Identification},
  author={Guo, Yu and Zhang, Xin and Wang, Chen and Li, Jiahui and Zhou, Peng},
  journal={Neural Networks},
  volume={180},
  pages={106--123},
  year={2025},
  publisher={Elsevier}
}

@article{blinowska2006electroencephalography,
  title={Electroencephalography (eeg)},
  author={Blinowska, Katarzyna and Durka, Piotr},
  journal={Wiley encyclopedia of biomedical engineering},
  volume={10},
  pages={9780471740360},
  year={2006},
  publisher={Wiley Hoboken, NJ, USA}
}

@article{sanchez2015should,
  title={Should epileptiform discharges be treated?},
  author={Sanchez Fernandez, Ivan and Loddenkemper, Tobias and Galanopoulou, Aristea S and Mosh{\'e}, Solomon L},
  journal={Epilepsia},
  volume={56},
  number={10},
  pages={1492--1504},
  year={2015},
  publisher={Wiley Online Library}
}

@article{hardmeier2014reproducibility,
  title={Reproducibility of functional connectivity and graph measures based on the phase lag index (PLI) and weighted phase lag index (wPLI) derived from high resolution EEG},
  author={Hardmeier, Martin and Hatz, Florian and Bousleiman, Habib and Schindler, Christian and Stam, Cornelis Jan and Fuhr, Peter},
  journal={PloS one},
  volume={9},
  number={10},
  pages={e108648},
  year={2014},
  publisher={Public Library of Science San Francisco, USA}
}

@article{obeid2016tuh,
  title={The Temple University Hospital EEG Data Corpus},
  author={Obeid, Iyad and Picone, Joseph},
  journal={Frontiers in Neuroscience},
  volume={10},
  pages={196},
  year={2016},
  publisher={Frontiers Media SA}
}
\endgroup

\end{document}